\documentclass[runningheads]{llncs}
\usepackage{xcolor}
\usepackage{graphicx} 
\usepackage{amsmath}
\usepackage{amsfonts}
\usepackage{amssymb}
\usepackage{amsthm}
\usepackage{bbm}
\usepackage{hyperref}
\newcommand{\Var}{\text{Var}}
\newcommand{\Cov}{\text{Cov}}

\newcommand{\model}{\text{model}}
\newcommand{\apx}{\text{approx}}
\newcommand{\CV}{\text{CV}}
\newcommand{\boot}{\text{boot}}
\newcommand{\Sc}{{S^C}}
\newcommand\argmin{\mathop{\mbox{{\rm argmin}}}\limits}

\newcommand\encircle[1]{%
  \tikz[baseline=(X.base)] 
    \node (X) [draw, shape=circle, inner sep=0] {\strut #1};}
\newcommand{\acirc}{\encircle{a}}
\newcommand{\bcirc}{\encircle{b}}
\newcommand{\ccirc}{\encircle{c}}
\usepackage{tikz}

\usepackage{subcaption}

\usepackage{chngcntr}
\begin{document}

\title{Stabilizing Estimates of Shapley Values with Control Variates}

\author{Jeremy Goldwasser\inst{1} \and
Giles Hooker\inst{2}}
\authorrunning{Goldwasser and Hooker}
%
\institute{UC Berkeley, Department of Statistics \\  367 Evans Hall, Berkeley, CA 94720 \\ \email{jeremy\_goldwasser@berkeley.edu} \and
University of Pennsylvania, Department of Statistics \& Data Science \\ 265 South 37th Street, Philadelphia PA 19104 \\ \email{ghooker@wharton.upenn.edu}}
%
\let\oldmaketitle\maketitle

\renewcommand{\maketitle}{\oldmaketitle\setcounter{footnote}{0}}

\maketitle

\begin{abstract}
    Shapley values are among the most popular tools for explaining predictions of black-box machine learning models. However, their high computational cost motivates the use of sampling approximations, inducing a considerable degree of uncertainty. To stabilize these model explanations, we propose ControlSHAP, an approach based on the Monte Carlo technique of control variates. Our methodology is applicable to any machine learning model and requires virtually no extra computation or modeling effort. On several high-dimensional datasets, we find it can produce dramatic reductions in the Monte Carlo variability of Shapley estimates.
\end{abstract}

\section{Introduction}
Machine learning models have become widely deployed in high-stakes domains such as healthcare, finance, and criminal justice \cite{ML_healthcare,ML_finance,ML_crime,ML_crime2}. Use of algorithmic decision-making in these areas has profound human consequences, but the predictive models themselves may be opaque. This motivates the development of methods to understand the reasoning behind the predictions of a black-box model \cite{XAI}. 

To that end, feature importance methods attribute a value to each feature passed into the model, indicating its predictive importance \cite{counterfactuals,SHAP,SAGE,LIME,RFs}. In this paper, we focus on Shapley values, a seminal concept in  game theory which have emerged as one of the most popular tools for model interpretability \cite{shapley_original,SHAP}. In layman's terms, Shapley values quantify how much information is gained from being told the value of each feature, averaged over orderings in which the features are revealed. 

Shapley values are rarely computed exactly, as the computational cost is exponential in the number of input features. Rather, they are typically estimated using the Shapley Sampling or KernelSHAP algorithm \cite{SHAP,Strumbelj2010,strumbelj2014}. These algorithms, however, are subject to sampling variability; as a result, running the same procedure twice may yield different estimated Shapley values, including different estimated orderings of features. This instability raises questions about the trustworthiness of insights gleaned from Shapley values.  

We seek to mitigate this issue by employing Monte Carlo variance reduction techniques. In particular, we use control variates, a method that adjusts one random estimator based on the known error of another. Here, the related estimator approximates the Shapley values of a first or second order Taylor expansion to the original model, depending on whether the value function assumes features are correlated or independent. In the independent case, these estimates entail essentially no additional computation; otherwise we must put some effort into pre-computing terms which can then be used for any query point for which we need to calculate Shapley values.  While variations on our methods are possible, from tuning parameters to more complex Monte Carlo schemes \cite{oates2017control}, our goal is to provide a default scheme that requires minimal computational or intellectual effort.  

Estimates of Shapley values are significantly more stable with control variates. In some cases, variance reductions reach as high as 90\% for dozens of features, along with noticeably more consistent rankings of Shapley values among features.\footnote{\url{https://github.com/jeremy-goldwasser/ControlSHAP} contains our code and experimental results.}

\section{Shapley Values in Machine Learning}
\subsection{Shapley Values}

In the context of machine learning, the Shapley values for an input sample $x \in \mathbb{R}^d$ present the change in prediction - i.e. the information gained - from learning the value of each feature. They do so by averaging the marginal contribution of a feature to every subset, or ``coalition," of other features.

Formally, let $f:\mathcal{X}\rightarrow\mathbb{R}$ be a machine learning model trained on $X \in \mathbb{R}^{n \times d}$. Further, let $S \subseteq [d] :=\{1, \ldots, d\} $ denote a subset of the $d$ features. Following SHAP \cite{SHAP}, the value function for input $x$ maps a subset $S$ to an estimate of the mean prediction, conditioned on the known features $x_S$: 

\begin{equation}\label{eq: val fn}
    v_x(S) := \hat{\mathbb{E}}[f(x_S, X_{[d]\backslash S})] = \hat{\mathbb{E}}[f(X) | X_S=x_S].
\end{equation}

In some settings - e.g. if the features are multivariate Gaussian or mutually independent - the conditional mean can be computed exactly. Otherwise, when the conditional distribution is intractable, the true mean is impossible to compute. Hence, in this case the value function $v_x(S)$ estimates ${\mathbb{E}}[f(x_S, X_{[d]\backslash S})]$ using an approximation of the conditional distribution $X|X_S$. 

The j\textsuperscript{th} Shapley value for input $x$, which we denote $\phi_j(x)$, averages the change in value function across all subsets of $[d]$, with and without the $j$\textsuperscript{th} feature.

\begin{equation}\label{Shapley Value}
    \phi_j(x) = \frac{1}{d} \sum_{S \subseteq [d]\backslash\{j\}} {d-1\choose |S|}^{-1} \big(v_x(S \cup \{j\}) - v_x(S)\big).
\end{equation}

Each Shapley value is a function of $\mathcal{O}(2^d)$ subsets. Since this is computationally prohibitive in high dimensions, we consider sampling approximations. 

\subsection{Shapley Estimation}

Shapley sampling \cite{Strumbelj2010} is the most straightforward algorithm to approximate Shapley Values. For each iteration $m \in \{1,\ldots,M\}$, randomly permute the $d$ features, and select the subset $S^m \subseteq [d]\backslash\{j\}$ that appears before $j$. Shapley sampling averages the change in value function from adding feature $j$ to each subset. 

\begin{equation}\label{Shapley Sampling}
    \hat{\phi}_j(x) = \frac{1}{M} \sum_{m=1}^M v_x(S^m \cup \{j\}) - v_x(S^m).
\end{equation}

In contrast, the KernelSHAP method \cite{SHAP} estimates all $d$ features at once. Again, in each iteration $m$ we select a coalition $S^m$ and compute its value function $v_x(S^m)$. Rather than sampling coalitions via permutations, however, $S^m$ is sampled from the distribution 

\begin{equation}\label{KSHAP kernel}
    p(S) \propto \frac{d-1}{{d\choose |S|}(|S|)(d-|S|)}.
\end{equation}

Each coalition $S$ can be equivalently expressed as a binary vector $z \in \{0, 1\}^d$ by defining $z_i = \mathbbm{1}\{i \in S\}$ for all $i \in [d]$. Accordingly, $z$ vectors $\mathbf{0}$ and $\mathbf{1}$ correspond to $S=\emptyset$ and $[d]$, respectively.

Abusing notation, let $v_x(z) := v_x(S)$, where $z$ is the binary representation of $S$. Consequently, we can express $v_x(\mathbf{0}) = \mathbb{E}[f(X)]$ and $v_x(\mathbf{1}) = f(x)$. The KernelSHAP estimator is defined as
\begin{align}\label{KernelSHAP}
    \hat{\phi}(x) & = \argmin_{\phi \in \mathbb{R}^d} \frac{1}{M} \sum_{m=1}^M \big(\phi^T z^m - (v_x(z^m)-v_x(\mathbf{0}))\big)^2  \nonumber \\
    & \qquad \text{s.t. \quad} \mathbf{1}^T\phi = v_x(\mathbf{1})-v_x(\mathbf{0}) \\
    & = (Z^TZ)^{-1} \left(I - \frac{ \mathbf{1}  \mathbf{1}^T (Z^TZ)^{-1}}{\mathbf{1}^T (Z^TZ)^{-1} \mathbf{1}} \right)Z^T (V-v_x(\mathbf{0})) \nonumber\\
    & \qquad \qquad \qquad \qquad  + \left( \frac{v_x(\mathbf{1})-v_x(\mathbf{0})}{\mathbf{1}^T (Z^TZ)^{-1} \mathbf{1}}\right) (Z^TZ)^{-1} \mathbf{1},  \nonumber
\end{align}

where $Z \in \mathbb{R}^{M \times d}$ and $V \in \mathbb{R}^M$ contain the coalitions $z^m$ and values $v_x(z^m)$. 

Drawing from \cite{kernel1988}, \cite{SHAP} proved that the weighting kernel in \ref{KSHAP kernel} produces linear regression coefficients that are equal to the Shapley values.

\subsection{Related Work}

A number of works seek to reduce the sample variability of Shapley values. Mitchell et al. propose several Monte Carlo techniques for choosing subsets \cite{SHAP_MC}; independently, Covert and Lee advocate antithetic sampling for KernelSHAP \cite{CovertLee}. Song et al. accelerate computation by storing pre-computed subsets \cite{SHAP_fast_sensitivity}. Our method can be applied on top of these methods, leveraging their contributions for added benefit.

Alternatives to sampling-based methods have also been proposed. Chen et al. derive expressions for Shapley values in settings for which the features contribute to predictions with a well-understood graphical structure \cite{martin_mike}. FastSHAP directly predicts Shapley values using a neural network \cite{fastSHAP}; it generalizes well when the neural network is trained on a large, diverse dataset. Other works more efficiently compute Shapley values given specific predictive models, e.g. Tree SHAP and Deep SHAP \cite{treeSHAP,SHAP}. These methods can quickly estimate Shapley values or obtain them exactly. However, they are limited to settings that satisfy their preconditions on model types, data structures, and number of inputs to explain. 

\section{Control Variates for Shapley Values}\label{sec:methods}

\subsection{Control Variates}\label{CV section}

The method of control variates is a variance reduction technique for Monte Carlo sampling problems. To reduce the error of an estimator, it leverages information about the known error of another.

Suppose $\hat{A}$ and $\hat{B}$ are unbiased estimators of $A^*$ and $B^*$, where only $B^*$ is known. Define the control variates estimator as

\begin{equation}\label{basic CV}
    \tilde{A} := \hat{A} - c(\hat{B} - B^*).
\end{equation}

For any constant $c$, $\tilde{A}$ is an unbiased estimator of $A^*$. Consequently, its mean squared error (MSE) is equivalent to its variance. The optimal $c$ is selected with the following lemma, using plug-in estimates of the variance and covariance.

\begin{lemma}\label{lemma CV}
Suppose $\hat{A}, \hat{B}$ are unbiased estimators and $\rho_{\hat{A}, \hat{B}} > 0$, where $\rho$ is Pearson's correlation coefficient. Then the control variates estimator (\ref{basic CV}) has minimal variance $\Var(\tilde{A}) = (1 - \rho^2_{\hat{A}, \hat{B}}) \Var(\hat{A})$ when $c^* = \frac{\Cov(\hat{A}, \hat{B})}{\Var(\hat{B})}$.
\end{lemma}

\subsection{ControlSHAP}

Our method, ControlSHAP, applies the method of control variates to the Shapley values for input $x$. The estimands $A^*$ and $B^*$ in \ref{CV section} are represented by $\phi_j^\model(x)$ and $\phi_j^\apx(x)$; these are the j\textsuperscript{th} Shapley values of $f$ and its Taylor approximation. In the following subsections, we will elaborate on the approximate model and its true Shapley values $\phi^\apx(x)$. 

Either Shapley sampling or KernelSHAP can be used to compute $\hat{\phi}_j^\model(x)$ and $\hat{\phi}_j^\apx(x)$. These are both appropriate for control variates because they are essentially unbiased for the true Shapley value (Eq. \ref{Shapley Value}), whose value function is the approximate conditional mean (Eq. \ref{eq: val fn}). Trivially, Shapley sampling (Eq. \ref{Shapley Sampling}) is unbiased, and \cite{CovertLee} showed that KernelSHAP has negligible bias (Eq. \ref{KernelSHAP}). Prior works have established that these methods may be biased for the Shapley value whose value function is the true conditional mean, $v_x(S) = \mathbb{E}[f(X)|X_s]$; however, by definition this Shapley value is not the estimand for this problem, as these means are intractable \cite{many_shap,SHAP_dep,SHAP_dep2}. By virtue of \eqref{basic CV}, control variate corrections to these estimands do not change their bias. 

Whichever method we choose, we must use the same subsets $S$ in computing $\hat{\phi}_j^\model(x)$ and $\hat{\phi}_j^\apx(x)$ to ensure they are correlated. Following Eq. \ref{basic CV} and Lemma \ref{lemma CV}, the ControlSHAP estimator is

\begin{align}\label{eqn:ControlSHAP}
\hat{\phi}_j^\CV(x) &= \hat{\phi}_j^\model(x) - \hat{\alpha} \big(\hat{\phi}_j^\apx(x) - \phi_j^\apx(x)\big)\\
\text{where }\hat{\alpha} &= \frac{\widehat{\Cov}(\hat{\phi}_j^\model(x), \hat{\phi}_j^\apx(x))}{\widehat{\Var}(\hat{\phi}_j^\apx(x))} \nonumber.
\end{align}

If $\alpha$ is obtained using the oracle covariance and variance, then this estimator reduces variance and MSE by a factor of $\rho^2(\hat{\phi}_j(x)^\model, \hat{\phi}_j(x)^\apx)$.


\subsection{Independent Features}

To compute Shapley values at $x$, we must estimate the conditional mean for each sampled subset, $\mathbb{E}[f(X) | X_S=x_S]$. To do so, we generate a number of samples from $X|X_S$, compute $f(X)$ on each, and take a sample mean. The conditional distribution $X | X_S$ may be difficult or impossible to characterize. Because of this, the standard approach popularized by SHAP samples the features in $S^C$ from their marginal distributions \cite{Strumbelj2010,SHAP}. Doing so makes the implicit assumption that the features are independent for the sake of computing $\mathbb{E}[f(X) | X_S=x_S]$. 

Consider the second-order Taylor approximation of $f$ around $x$. If the value function is computed sampling features from their marginal distributions, then the Shapley values $\phi^\apx(x)$ of this approximate model can be computed exactly. While we refer to this as the ``independent features case," the following theorem holds \textit{regardless} of whether the features are actually independent.

\begin{theorem}\label{thm:quadratic shapley}
Define $\mu := \mathbb{E}[X]$ and $\Sigma_{jk} := Cov(X_j, X_k)$. Let the value function $v_x(S)$ compute conditional means by sampling each feature from its marginal distribution. The j\textsuperscript{th} Shapley value of the second-order Taylor approximation of $f$ around $x$ is 
\begin{align*}
    \phi_j^\apx(x) &= \frac{\partial f}{\partial x_j}(x_j - \mu_j)\\
    &\qquad- \frac{1}{2}\Bigg[\sum_{k=1}^d (x_k - \mu_k) \frac{\partial^2 f}{\partial x_j x_k}\Bigg] (x_j - \mu_j) \\
    &\qquad- \frac{1}{2} \sum_{k=1}^d \Sigma_{jk} \frac{\partial^2 f}{\partial x_j x_k}.
\end{align*}
\end{theorem}

Proof of Theorem \ref{thm:quadratic shapley} is in the section 1.1 of the appendix. 

\subsection{Correlated Features}

Input features are often correlated with one another. When this is the case, it may be imprudent to falsely assume independence for the sake of computational convenience. This may produce Shapley values that misrepresent the relationship between inputs and predictions, and rely on the values that a machine learning function makes in areas where it has very little data \cite{hooker2021unrestricted}.  

Unfortunately, respecting feature correlations requires that we sample from $X|X_S$ for all sampled subsets $S$. Nonparametric density estimation in high dimensions is challenging due to the curse of dimensionality, and most approaches typically require considerable modeling effort. 

To circumvent this, a common alternative estimates the conditional mean by characterizing $X|X_S$ as a multivariate normal distribution\cite{SHAP_dep,SHAP_dep2}. This makes the value function easy to compute, without having to ignore dependencies between the features. One can even faithfully represent binary data under this categorization by simulating Gaussian random variables and thresholding accordingly \cite{bindata}. 

Moreover, using this value function enables us to compute the true Shapley value of a first-order Taylor expansion of $f$. As in Theorem \ref{thm:quadratic shapley}, the formula in Theorem \ref{thm:depfeatures} holds whether or not the features are indeed normally distributed. 

\begin{theorem} \label{thm:depfeatures}
    Define the value function to take the conditional expectation of $f(X)|X_S$ over a multivariate normal distribution. Let $S^m \subseteq [d]\backslash\{j\}$ be the subset of features appearing before $j$ in the $m$\textsuperscript{th} permutation of $[d]$, and let $Q_S$ and $R_S$ be matrices defined in Section 1.2 of the supplementary material that do not depend on $x$. Define
    \begin{equation*}
        D_j := \frac{1}{d!} \sum_{m=1}^{d!} ([Q_{S_j^m \cup j} + R_{S_j^m \cup j}] - [Q_{S_j^m} + R_{S_j^m}]).
    \end{equation*}
    The j\textsuperscript{th} Shapley value of the first-order Taylor approximation of $f$ around $x$ is
    \begin{equation*}
        \phi_j^\apx(x) = \nabla f(x)^T D_j (x - \mu).
    \end{equation*}
\end{theorem}

Evaluating $\phi_j^\apx(x)$ still requires computing an exponential number of terms for $D_j$. However, because $D_j$ does not depend on $x$, we can reuse it for other inputs $x$. That is, we can front-load computational work into accurately estimating $D_j$; then, we can instantly obtain close approximations to $\phi_j^\apx(x)$ for as many inputs $x$ as we want. This is valuable if we want to estimate Shapley values at many inputs, either for future data or to provide a global notion of feature importance.

The normality condition in Therorem \ref{thm:depfeatures} can be relaxed somewhat. It is needed insofar as it produces conditional means of the form
\begin{equation*}
    E[X_1 | X_2 = x_2] = \mu_1 + \Sigma_{12}\Sigma_{22}^{-1}(x_2 - \mu_2),
\end{equation*}

\noindent which occur whenever the variance of the conditional distribution is independent of the values of the conditioning set. This property holds for all elliptically-symmetric distributions, including the multivariate $t$-, Laplace, and logistic distributions \cite{cond_mean}. 


\subsection{Non-Differentiable Models}

The discussion so far assumes that we have access to first and sometimes second derivatives of $f$. This excludes models based on trees that result in piecewise constant functions, as well as neural networks with ReLU activation functions, which have first but not second derivatives. While there are specific Shapley methods for trees \cite{campbell2022exact}, these only apply to models with axis-oriented splits (counter to e.g. \cite{tomita2020sparse}). 

For our purposes, we use a Taylor series only to provide linear and quadratic approximations to $f$.  Thus, we can replace $\partial f/\partial x_i$ and $\partial^2 f/\partial x_i \partial x_j$ with finite-difference approximations based on large step sizes, so as to approximate model behavior across substantial changes in the features.  

In our experiments below, we used finite differencing on random forest predictors. We chose the step size of each numeric feature to be its marginal standard deviation; for categorical features, we swapped the observed values. Alternative choices would include basing step sizes on conditional distributions, or tuning them for agreement with $f$. Our goal has been to examine a straightforward control variate method with minimal costs both computationally and in human effort, so we have not pursued these variations here. We speculate that additional tuning would provide further improvements, which may be worthwhile in particular applications.

\section{Variance Estimation}

To use the method of control variates, we need to estimate the variance and covariance of our two estimators, as in Eq. \ref{eqn:ControlSHAP}. This section outlines strategies for doing so. 

\subsection{Shapley Sampling}

Let $G_S := v_x(S\cup j) - v_x(S)$ for subset $S$. Shapley sampling selects $M$ subsets $S$ via independent permutations, then takes a sample mean of their $G_S$ values. Therefore the variance of Shapley sampling values is
\begin{equation*}
    \Var(\hat{\phi}_j) = \frac{1}{M} \Var(G_S).
\end{equation*}

We estimate $\Var(G_S)$ empirically across the $M$ observations of $G_S$. Estimating the covariance between the Shapley estimates is similarly straightfoward:


\begin{equation*}
    \Cov(\hat{\phi}_j^{\model}, \hat{\phi}_j^{\apx}) = \frac{1}{M} \Cov(G_S^\model, G_S^\apx),
\end{equation*}

\noindent which again we estimate empirically. 

\subsection{KernelSHAP}

We present three approaches to estimate the KernelSHAP variance and covariance. These are unrelated to the choice of independent or correlated features.
\begin{enumerate}
    \item \textbf{Bootstrapping}. For many iterations, take $M$ samples with replacement from the observed subsets $S$. Compute the KernelSHAP estimates $\hat{\phi}_\boot^{\model}$ and $\hat{\phi}_\boot^{\apx}$ on the pairs $\{(S_\boot, v_\boot^\model(S))\}_{i=1}^M$ and $\{(S_\boot, v_\boot^\apx(S))\}_{i=1}^M$. Repeat this process to produce many bootstrapped estimates of $\hat{\phi}_\boot^{\model}$ and $\hat{\phi}_\boot^{\apx}$, with which we compute empirical variance and covariance. 
    \item \textbf{Least Squares}. KernelSHAP learns $\hat{\phi}$ by fitting a constrained least squares (Eq. \ref{KernelSHAP}). 
    Writing $A(Z) =  (Z^TZ)^{-1} \left[I - (\mathbf{1} \mathbf{1}^T (Z^TZ)^{-1})/(\mathbf{1}^T (Z^TZ)^{-1} \mathbf{1}) \right]Z^T$ we make the approximation
    \begin{equation*}
        \widehat{\Var}(\hat{\phi}) = A(Z) \Var(v_x(S)) A(Z)^T.
    \end{equation*}
    The diagonals of $Var(v_x(S))$ are computed via sample variance, provided more than one sample is used to estimate each $v_x(S)$. Analogously,
    \begin{equation*}
        \widehat{\Cov}(\hat{\phi}^\model, \hat{\phi}^\apx) = A(Z) \widetilde{\Sigma} A(Z)^T\\
    \end{equation*}
    where $\widetilde{\Sigma} = \Cov(v_x(S)^\model, v_x(S)^\apx)$.
    
    Here we note that this formula ostensibly accounts only for the sampling variability in $v_x(S)$ rather than the Monte Carlo choice of coalitions $Z$. However, by the total covariance formula writing $\hat{\phi}= B(Z,V)$ we have that 
    \begin{align*}
     \Cov(&\hat{\phi}^\model, \hat{\phi}^\apx) \\
     &= \mathbb{E}_Z \Cov( B(Z,V^\model), B(Z,V^\apx)|Z) \\
     &\;\;\;+ \Cov_Z( \mathbb{E}_V B(Z,V^\model), \mathbb{E}_V B(Z,V^ \apx))
    \end{align*}
    \cite{CovertLee} demonstrated that kernelSHAP has small bias, making the second term small, and this calculation provides a consistent estimate of the first term. $\Var(\hat{\phi}^\model)$ is estimated with an equivalent calculation. 

    \item \textbf{Grouped}. \cite{CovertLee} introduced a method based on their finding that the variance evolves at a $O(\frac{1}{M})$ rate. First, split the $\{(S, v_x(S))\}_{i=1}^M$ pairs into disjoint groups, and fit $\hat{\phi}^{(k)}$ in each. Then, take the empirical variance and covariance of these $\hat{\phi}^{(k)}$ values, and scale linearly.  
\end{enumerate}

Empirically, we observed that the bootstrapped and least squares methods yield roughly the same variances and covariances. Section 4 of the appendix shows that ControlSHAP with these two methods has nearly identical performance, across multiple datasets and machine learning models. Our experimental results all use the least squares method, since bootstrapping has a moderately higher computational cost.


In our experiments, we were not able to get the grouped method to reliably work. We observed that it occasionally produced highly erroneous estimates of variance and covariance. While this may be remedied by fitting a larger number of groups, the group size must be large enough to avoid singularity issues in fitting $\hat{\phi}^{(k)}$. More details are in Section 4 of the appendix.

\section{Experiments}

We demonstrated the efficacy of our ControlSHAP methodology on five datasets. A simulated dataset contained ten normally distributed features with block covariance structure; each feature had a correlation of 0.5 with one feature and 0.25 with another, and the response given by a logistic regression. We also used four public datasets from the UCI ML repository: Portuguese bank \cite{bank}, German credit \cite{credit}, census income \cite{census}, and breast cancer subtyping \cite{brca}. 

These real-world datasets are higher-dimensional, containing roughly 20 features on average. Many features are categorical with more than two levels, e.g. Month. We represent each of these features as a set of one-hot encodings. These transformations are necessary to use models like logistic regression and neural networks; they are also required for ControlSHAP, which takes gradients with respect to the features. The Shapley value of a categorical feature is the sum across the individual levels \cite{SHAP}. 

In the following analyses, we refer to the two approaches for the value function as assuming features are independent versus correlated (dependent). This terminology only serves to distinguish the method of sampling and Taylor expansion. It does not entail making actual assumptions about the data, as detailed in Section \ref{sec:methods}. The ControlSHAP estimator's properties such as unbiasedness do not require feature independence or Gaussianity to hold. 

For each data set, and each model, we sampled 40 data points from the test set at random and estimated Shapley values 50 times for each, with and without control variates and using both correlated and independent sampling. For each estimate, we sampled 1000 coalitions. For KernelSHAP, we generated 10 points per coalition to estimate pointwise variances. We ran 200 bootstrap samples and 20 groups to estimate variances and covariances for control variates for bootstrap and grouped estimators. For Shapley sampling, which re-runs the Monte Carlo calculation for each feature, we used 1 sample per coalition.  Computations were run on an internal cluster.

\begin{figure}
\centering\includegraphics[width=0.7\textwidth]{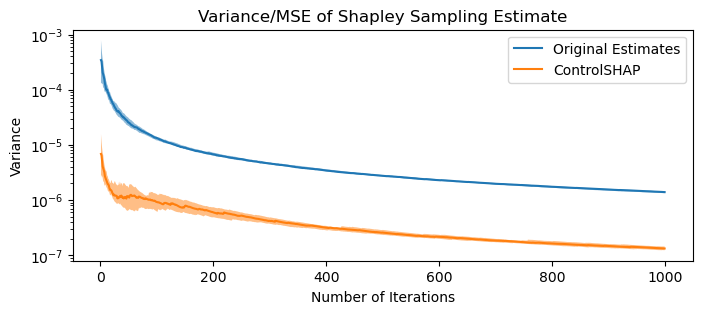}
\caption{Variance (or MSE) of Shapley estimates. Computed independent-features Shapley sampling estimates on ``Prev Days" feature, using logistic regression model on Bank dataset. Bands show inner quartiles across 50 iterations.}
\label{fig:Var by perms}
\end{figure}

As Figure \ref{fig:Var by perms} demonstrates, ControlSHAP significantly improves the variability of Shapley estimates, regardless of the number of samples drawn. For this feature - chosen because it has the highest Shapley value - adjusting the Shapley estimate consistently reduces its variance by at least 90\%. 

Next, we consider the improvements in variance - equivalently, in MSE - across all 5 datasets, using a variety of machine learning predictors. We calculated the variance reduction of our ControlSHAP methods by taking the sample variance of each feature's Shapley value across 50 iterations.

\begin{equation}\label{eqn:var reduc}
    \text{VarReduc}_j(x) = \frac{\Big[\widehat{\Var}(\hat{\phi}_j^\model (x)) - \widehat{\Var}(\hat{\phi}_j^\CV(x))\Big]}{\widehat{\Var}(\hat{\phi}_j^\model(x))}.
\end{equation}

\begin{figure}
\centering\includegraphics[width=0.8\textwidth]{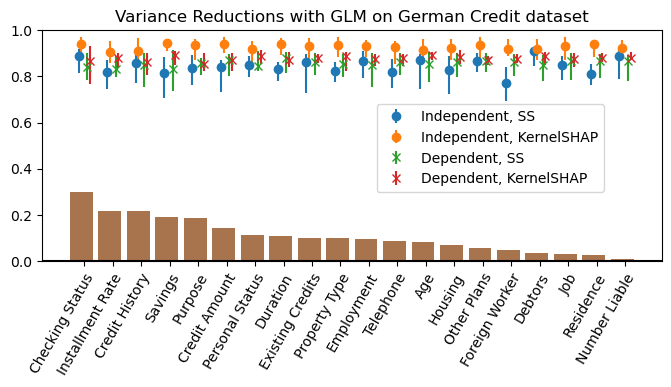}
\caption{Variance reductions of ControlSHAP methods. Sorted features by absolute Shapley values (unscaled in plot) with logistic regression predictor. Confidence bands denote lower and upper quartiles of Eq. \ref{eqn:var reduc} across 40 held-out values of $x$.}
\label{fig:Var reducs, Bank}
\end{figure}

Figure \ref{fig:Var reducs, Bank} displays results for all features of the German Credit dataset, using logistic regression as a predictor. Our four ControlSHAP methods produce large variance reductions on all features, whose Shapley values span several orders of magnitude. On average, our final Shapley estimates have roughly 80-95\% lower variability than the original estimates.

A second stability metric we used was the number of ranking changes. This metric is motivated by the fact that often practitioners care more about identifying features with high Shapley values than the values themselves \cite{top_k_shapley_neuron,top_k_shapley_graph,top_k_shapley_management}. Formally, let $\hat{r}_j(x)$ be the ranking of feature $j$ within a set of $d$ Shapley estimates. For each method, we compute the average number of pairwise ranking changes across 50 resamplings. 

\begin{equation}\label{eqn: rank chgs}
    \text{RankChgs}(x) = {50\choose2}^{-1}\sum_{a \neq b} \Big[\sum_{j=1}^d |\hat{r}_j^a(x) - \hat{r}_j^b(x)|\Big].
\end{equation}

\begin{figure}
\centering\includegraphics[width=0.7\textwidth]{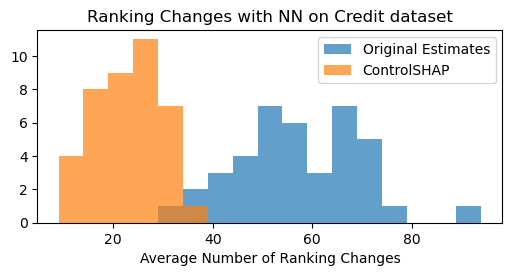}
\caption{Average number of Ranking Changes (Eq. \ref{eqn: rank chgs}) across 50 iterations, with and without ControlSHAP correction. Shapley values estimated on 40 inputs via KernelSHAP with correlated features. Neural network trained on Credit dataset.} 
\label{fig:Ranking Changes}
\end{figure}

Figure \ref{fig:Ranking Changes} shows the average number of ranking changes for 40 sample inputs. We observe that the rankings are significantly more stable with the control variate. ControlSHAP averages around 20 rank changes, compared to 60 for the original estimates. Note that this statistic highly depends on the stability of Shapley sampling prior to using control variates: if the original estimates exhibit no ranking changes between simulations, this statistic will not show an improvement even if they variance of the estimates is reduced. We believe our simulations to be representative of current practice where a noticeable reduction is evident. 

\begin{table*}[h]\label{tbl: Results}
\caption{ControlSHAP Percent Reductions in Variance (left) and Ranking Changes (right). Shapley Sampling (SS) and KernelSHAP (KS), with and without independence assumption. Averaging variance and rank-change reductions over 40 sample inputs. Taking median variance reduction of features with 5 highest Shapley values.}

\begin{center}
\begin{tabular}{||c||c|c|c|c|c|}
\hline
\textbf{\;Logistic Reg.} \;&\; \textbf{SIM} \;&\; \textbf{BANK} \;&\; \textbf{BRCA} \;&\; \textbf{CENSUS} \;&\; \textbf{CREDIT}\; \\
\hline
\textit{\;Indep., SS} \;&\;10\% $\vert$ 6\% \;&\;76\% $\vert$ 57\% \;&\;75\% $\vert$ 57\% \;&\;33\% $\vert$ 13\% \;&\;83\% $\vert$ 60\% \\
\hline
\textit{\;Indep., KS} \;&\;24\% $\vert$ 17\% \;&\;84\% $\vert$ 43\% \;&\;87\% $\vert$ 54\% \;&\;3\%  $\vert$ 2\% \;&\;94\% $\vert$ 67\% \\
\hline
\textit{Correlated, SS}   \;&\;59\% $\vert$ 32\% \;&\;67\% $\vert$ 42\% \;&\;92\% $\vert$ 68\% \;&\;71\% $\vert$ 58\% \;&\;85\% $\vert$ 64\% \\
\hline
\textit{Correlated, KS} \;&\;51\% $\vert$ 27\% \;&\;69\% $\vert$ 29\% \;&\;94\% $\vert$ 71\% \;&\;59\% $\vert$ 30\% \;&\;87\% $\vert$ 59\% \\
\hline
\end{tabular}\label{tbl: Results}
\end{center}

\begin{center}
\begin{tabular}{||c||c|c|c|c|c|}
\hline
\textbf{\;Neural Net} &\; \textbf{SIM} \;&\; \textbf{BANK} \;&\; \textbf{BRCA} \;&\; \textbf{CENSUS} \;&\; \textbf{CREDIT}\; \\

\hline
\textit{\;Indep., SS} \;&\;9\% $\vert$ 6\% \;&\;8\% $\vert$ 2\% \;&\;76\% $\vert$ 52\% \;&\;4\% $\vert$ 3\% \;&\;55\% $\vert$ 36\% \\ 
\hline
\textit{\;Indep., KS} \;&\;22\% $\vert$ 15\% \;&\;25\% $\vert$ 8\% \;&\;86\% $\vert$ 55\% \;&-1\%$\vert$ 1\% \;&\;80\% $\vert$ 45\% \\
\hline
\textit{Correlated, SS} \;&\;58\% $\vert$ 31\% \;&\;58\% $\vert$ 37\% \;&\;92\% $\vert$ 65\% \;&\;61\% $\vert$ 42\% \;&\;84\% $\vert$ 61\% \\ \hline
\textit{Correlated, KS} \;&\;51\% $\vert$ 28\% \;&\;40\% $\vert$ 13\% \;&\;92\% $\vert$ 68\% \;&\;51\% $\vert$ 24\% \;&\;87\% $\vert$ 60\% \\
\hline
\end{tabular}
\end{center}

\begin{center}
\begin{tabular}{||c||c|c|c|c|c|}
\hline
\textbf{\;RF} &\; \textbf{SIM} \;&\; \textbf{BANK} \;&\; \textbf{BRCA} \;&\; \textbf{CENSUS} \;&\; \textbf{CREDIT}\; \\
\hline 
\textit{\;Indep., SS} \;&\;31\% $\vert$ 15 \;&\;18\% $\vert$ 3\% \;&\;19\% $\vert$ 7\% \;&\;4\%$\vert$ 2\% \;&\;15\% $\vert$ 6\% \\
\hline 
\textit{\;Indep., KS} \;&\;37\% $\vert$ 19\% \;&\quad4\%  $\vert$ 3\% \;&\;14\% $\vert$ 6\% \;&7\%$\vert$ 4\% \;&\;14\% $\vert$ 6\% \\
\hline 
\textit{Correlated, SS} \;&\;56\% $\vert$ 31\% \;&\quad6\%  $\vert$ 5\% \;&\;22\% $\vert$ 11\% \;&\;2\% $\vert$ 5\% \;&\;24\% $\vert$ 17\% \\
\hline 
\textit{Correlated, KS} \;&\;54\% $\vert$ 31\% \;&\quad1\%  $\vert$ -2\% \;&\;17\% $\vert$ 9\% \;&\;3\% $\vert$ -1\% \;&\;22\% $\vert$ 10\% \\
\hline 
\end{tabular}
\end{center}

\end{table*}

Table \ref{tbl: Results} contains results across all 5 datasets with logistic regression, neural network, and random forest predictors. In addition to reductions in variance, we also present reductions in the number of rank changes, averaged over 40 sample inputs. Full sets of visualizations for variance reductions and rank changes can be found in Section 2 of the appendix.

In every setting, ControlSHAP reduces the variability of Shapley estimates, often by a wide margin. For both logistic regression and neural networks, at least one ControlSHAP method produces over a 50\% reduction in variance and 30\% reduction in number of rank changes on all five datasets. On the BRCA and German credit datasets, these figures rise to 85\% and 55\% for most ControlSHAP methods. Every ControlSHAP method reduces variance by over 67\% on three of the four real-world datasets with logistic regression.

While ControlSHAP is generally strong for neural networks, it tends to work better when in the dependent features case. On the simulated, bank, and census datasets, variance reductions are typically below 25\% assuming independence and above 50\% otherwise. Presumably, this owes to the fact that neural networks are less smooth. The quadratic model from the independent features case (Thm. \ref{thm:quadratic shapley}) is more prone to neural networks' steep gradients and hessians, and may poorly approximate model behavior as a result.

When the machine learning model is not differentiable, gradients and hessians can be taken via finite differencing. These coarse estimates use a smooth approximation to step-wise constant model behavior, which also results in a lower correlation between our target and control variate. 

\begin{figure}[h]
\centering\includegraphics[width=0.5\textwidth]{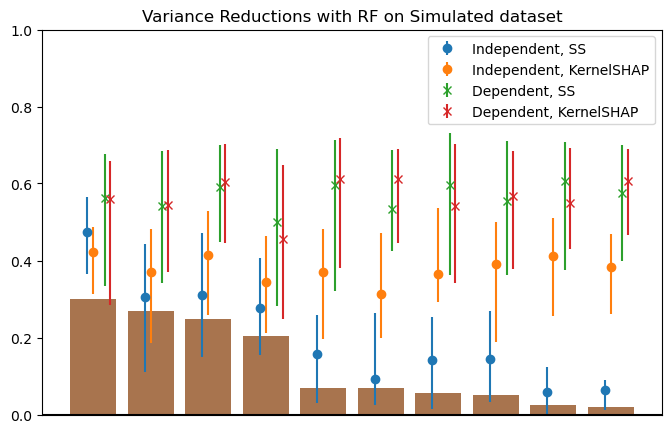}
\caption{Variance Reductions of Random Forest on simulated dataset with 10 features. Upper and lower quantiles of reductions provided across 40 inputs. Computed gradients via finite differencing.}
\label{fig:RF}
\end{figure}

Nevertheless, ControlSHAP increases stability on random forests, again particularly in the correlated features case. Figure \ref{fig:RF} showcases its solid performance on the simulated dataset, reducing variability by over 50\% with dependent features. Table \ref{tbl: Results} shows its reductions in variance and rank changes on all datasets and ControlSHAP methods. 

\begin{figure}
\centering\includegraphics[width=0.7\textwidth]{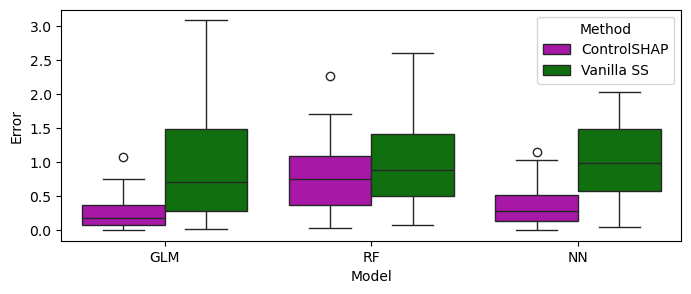}
\caption{Normalized absolute difference from sum of Shapely estimates to $f(x)-Ef(X)$. Ran  for 10 iterations on credit dataset with independent features.}
\label{fig:unbiased}
\end{figure}

Finally, we analyzed the ControlSHAP estimates to confirm they are unbiased. While the true SHAP values are computationally prohibitive to compute, their sum is known to be $f(x)-E[f(X)]$ by the efficiency property\cite{SHAP,shapley_original}. To search for bias, we compared the sum of Shapley estimates from this target, across multiple ML predictors. Figure \ref{fig:unbiased} demonstrates that ControlSHAP exhibits no evidence of bias; on the contrary, the sum of its Shapley estimates is typically closer to the target. 

\section{Discussion}

While Shapley values are capable of providing useful model explanations, they fall short in one critical way: stability. Stability is necessary to instill trust in our interpretations of black-box machine learning algorithms. Practitioners cannot use such a system with confidence if rerunning the same procedure yields different explanations. 

This work proposes a method, ControlSHAP, to address this. ControlSHAP can be applied in conjunction with other attempts to mitigate sampling variability. Further, while designed for differentiable models, it can help explain other predictive models via finite differencing. Specifically, it adjusts Shapley estimates using the method of control variates, which track the extent to which the random samples over- or underestimate on a correlated estimator. 

ControlSHAP stabilizes Shapley estimates across a wide range of datasets and algorithms, in some cases reducing variability by as high as 90\%. Moreover, we can easily obtain accurate estimates of the added benefit from performing ControlSHAP adjustments. Recall that ControlSHAP reduces variance by approximately $\rho^2(\hat{\phi}_j(x)^\model, \hat{\phi}_j(x)^\apx)$ (Eq. \ref{eqn:ControlSHAP}). We can estimate this correlation using the variance and covariance used to compute $\widehat{\alpha}$; this enables us to estimate the variance reduction. As section 3 in the appendix demonstrates, these anticipated reductions closely mirror the empirical quantities. 

Alternatively, ControlSHAP can be used to accelerate convergence. \cite{CovertLee}, for example, suggest terminating when the variability falls below a certain threshold. Using the control variate here could lead to a dramatic speed-up. Note in Figure \ref{fig:Var by perms} that the ControlSHAP estimates have comparable variability after 100 iterations as the original values after 1000. 

ControlSHAP can be employed as a relatively ``off-the-shelf" tool, in the sense that it stabilizes Shapley estimates with close to no extra computational or modeling work. The only model insight required is the gradient, as well as the hessian in the independent features case. Computationally, the single substantial cost is in the correlated features case, which requires accurate estimation of pre-compute matrices. Otherwise, ControlSHAP only necessitates passing each $X|X_S$ through the Taylor approximation, and computing the Shapley values' covariance - both of which are extremely quick tasks. 

It is worth noting that other explanation algorithms also suffer from various forms of instability. \cite{S-LIME} and \cite{distillation_trees} found that the explanations from LIME and distillation trees are sensitive to the synthetic data used to generate them. In addition, counterfactual explanations for adjacent points may be subject to vary widely \cite{counterfactual_discussion}. 

These drawbacks motivate future work stabilizing model explanations. Shapley computations may benefit from tuning finite differences, more sophisticated Monte Carlo schemes, or alternative control variates.  In some contexts it may be preferable to use inherently interpretable models like decision trees or the LASSO \cite{cynthia_rudin_stop,LASSO}. We intend ControlSHAP to complement these models, enabling more trustworthy use of black-box methods. 

\bibliographystyle{splncs04}
\bibliography{refs}
\setcounter{section}{0}

\section*{Appendix}
\section{Exact Shapley Values}

\subsection{Shapley Value, Assuming Feature Independence}
\begin{proof}

Assume features are sampled independently in the value function. Consider a second-order Taylor approximation to $f$ at $x$:

\begin{equation*}
    g(x') := f(x) + (x'-x)^T \nabla f(x) + \frac{1}{2}(x'-x)^T \nabla^2 f(x) (x'-x).
\end{equation*}

Let $J = \nabla f(x)$, $H = \nabla^2 f(x)$, $\Sigma=\Cov(X)$, and $\sigma^2_{k\ell}=\Cov(X_k, X_\ell)$. The Shapley value function is
\begin{align*}
    v_x(S) &= E[g(X')|X'_S = x_S]\\ 
    &= f(x) + E[(x'-x)|X'_S=x_S]^T \nabla f(x) \\
    &\qquad\qquad + \frac{1}{2}E\Big[(x'-x)^T \nabla^2 f(x) (x'-x)|X'_S=x_S\Big]\\
    &= f(x) + (\mu_{S^C} - x_{S^C})^T J_{S^C}\\
    &\qquad\qquad + \frac{1}{2}E\Big[(x_\Sc'-x_\Sc)^T \nabla^2 f(x)_{\Sc\Sc} (x_\Sc'-x_\Sc)\Big],
\end{align*}

where the quadratic term is equal to

\begin{align*}
  &tr(H_{\Sc\Sc}\Sigma_{\Sc\Sc}) 
    + (\mu_\Sc - x_\Sc)^T H_{\Sc\Sc}(\mu_\Sc - x_\Sc)\\
    = &\sum_{k\in \Sc}\sum_{\ell\in \Sc} 
    H_{k\ell}\Big(\sigma^2_{k\ell} + (\mu_k - x_k)(\mu_\ell - x_\ell)\Big)\\
    = &H_{jj}(\sigma^2_{jj} + (\mu_j-x_j)^2).
\end{align*}

Recall $\displaystyle\phi_j(x) := \frac{1}{d} \sum_{S \subseteq [d]\backslash\{j\}} {d-1\choose |S|}^{-1} \big(v_x(S \cup \{j\}) - v_x(S)\big)$. The difference in value functions is

\begin{align*}
    v_x(S\cup j)-v_x(S) &= \underbrace{(x_j-\mu_j) J_j - \frac{1}{2}H_jj\Big(\sigma^2_j+(\mu_j-x_j)^2\Big)}_{\acirc}\\
    &\qquad\qquad - \underbrace{\sum_{k\in (S\cup j)^C} \underbrace{H_{jk}\Big[\sigma^2_{jk} + (\mu_j-x_j)(\mu_k-x_k))\Big]}_{\ccirc}}_{\bcirc}.
\end{align*}

Define the Shapley weight $w_S = \frac{1}{d}{d-1\choose |S|}^{-1}$.

\begin{align}\label{quadratic shapley weighted}
\begin{split}
    \phi_j(x) &= \sum_{S \subseteq [d]\backslash\{j\}}w_S(\acirc - \bcirc)\\
    &= \acirc \sum_{S \subseteq [d]\backslash\{j\}} w_S - \sum_{S \subseteq [d]\backslash\{j\}} w_S \bcirc\\
    \sum_{S \subseteq [d]\backslash\{j\}} w_S \bcirc &= \sum_{S \subseteq [d]\backslash\{j\}} w_S \sum_{k\in \{S\cup j\}^C} \ccirc\\
    &= \sum_{S \subseteq [d]\backslash\{j\}} \sum_{k\in \{S\cup j\}^C} w_S \ccirc\\
    &= \sum_{k\neq j} \sum_{S \subseteq [d]\backslash\{j,k\}} w_S \ccirc\\
    &= \sum_{k\neq j} \ccirc \sum_{S \subseteq [d]\backslash\{j,k\}} w_S.
\end{split}
\end{align}

Noting subsets of equal size have the same Shapley weight, we can easily show $\displaystyle \sum_{S \subseteq [d]\backslash\{j\}}w_S = 1$. 

\begin{align}\label{sum all weights}
\begin{split}
    \sum_{S \subseteq [d]\backslash\{j\}}w_S &= \sum_{S \subseteq [d]\backslash\{j\}} \frac{1}{d}{d-1\choose |S|}^{-1}\\ 
    &= \sum_{a=0}^{d-1} {d-1\choose a} \frac{1}{d}{d-1\choose a}^{-1} = \sum_{a=0}^{d-1}\frac{1}{d}=1
\end{split}
\end{align}

With a bit more arithmetic, we can show $\displaystyle \sum_{S \subseteq [d]\backslash\{j, k\}}w_S = \frac{1}{2}$.

\begin{align}\label{sum half weights}
\begin{split}
    \sum_{S \subseteq [d]\backslash\{j, k\}}w_S &= \sum_{a=0}^{d-2} {d-2\choose a} \frac{1}{d}{d-1\choose a}^{-1}\\
    &= \sum_{a=0}^{d-2} \frac{(d-2)!}{a!(d-a-2)!}\frac{a!(d-a-1)!}{d!}\\
    &= \sum_{a=0}^{d-2} \frac{d-a-1}{d(d-1)}\\
    &= \frac{1}{d(d-1)}\Big[d(d-1) - \sum_{a=0}^{d-2}(a+1) \Big]\\
    &= \frac{1}{d(d-1)}\Big[d(d-1) - \frac{d(d-1)}{2} \Big] = \frac{1}{2}.
\end{split}
\end{align}

Plugging the results of \ref{sum all weights} and \ref{sum half weights} into \ref{quadratic shapley weighted} yields the final expression for the Shapley value:

\begin{align*}
    \phi_j(x) &= \acirc \sum_{S \subseteq [d]\backslash\{j\}} w_S - \sum_{S \subseteq [d]\backslash\{j\}} w_S \bcirc\\
    &= \acirc - \frac{1}{2} \sum_{k\neq j}\ccirc\\
    &= (x_j-\mu_j) J_j - \frac{1}{2}H_jj\Big(\sigma^2_j+(\mu_j-x_j)\Big)\\
    &\qquad\qquad - \frac{1}{2} \sum_{k\neq j} H_{jk}\Big[\sigma^2_{jk} + (\mu_j-x_j)(\mu_k-x_k))\Big]\\
    &= (x_j-\mu_j) J_j - \frac{1}{2} \Big[\sum_{k=1}^d (x_k-\mu_k) H_{jk})\Big](x_j-\mu_j)\\
    &\qquad\qquad - \frac{1}{2}\sum_{k=1}^d \sigma^2_{jk}H_{jk}.
\end{align*}
\end{proof}

\subsection{Shapley Value, Correlated Features}

Consider the linear model $g(x') = \beta^T x' + b$, where $x' \sim \mathcal{N}(\mu, \Sigma)$. (For our problem, $\beta = \nabla f(x)$ and $b = f(x)$.) 

Let $P_S$ be the projection matrix selecting set $S$; define $R_S = P_{\bar S}^T P_{\bar S} \Sigma P_S^T (P_S \Sigma P_S^T)^{-1} P_S$ and $Q_S = P_S^T P_S$. We consider the set of permutations of $[d]$, each of which indexes a subset $S$ as the features that appear before $j$. Averaging over all such permutations, the Shapley values of $g(x)$ \cite{SHAP,SHAP_dep,SHAP_dep2} are

\begin{align*}
    \phi_j(x') &= \beta \underbrace{\left [\frac{1}{d!} \sum_m ([Q_{\{S^m \cup j\}^C} - R_{S^m \cup j}]- [Q_{\{S^m\}^C} - R_{S^m}])\right ]}_{C_j}   \mu\\ 
    &\qquad\qquad + \beta \underbrace{\left [ \frac{1}{d!} \sum_m ([Q_{S^m \cup j} + R_{S^m \cup j}] - [Q_{S^m} + R_{S^m}]) \right ]}_{D_j} x'.
\end{align*}

Lastly, we observe that $C_j = -D_j$. For each subset $S$, the $R$ terms cancel out; $Q_{S^C} + Q_S = I_d$, so $Q_{\{S\cup j\}^C} - Q_{S^C} = -(Q_{S\cup j} - Q_S)$. This yields our expression for the dependent Shapley value:

\begin{equation*}
    \phi_j^g(x) = \beta^T D_j (x - \mu).
\end{equation*}

\section{Comprehensive Results}

 We display results across all 5 datasets and 3 machine learning predictors. All datasets were binary classification problems, so we used the same models to fit them. For logistic regression and random forest, we used the sklearn implementation with default hyperparameters. For the neural network, we fit a two-layer MLP in Pytorch with 50 neurons in the hidden layer and hyperbolic tangent activation functions. 
 
\begin{figure}
     \begin{subfigure}[b]{\textwidth}
         \includegraphics[width=\textwidth]{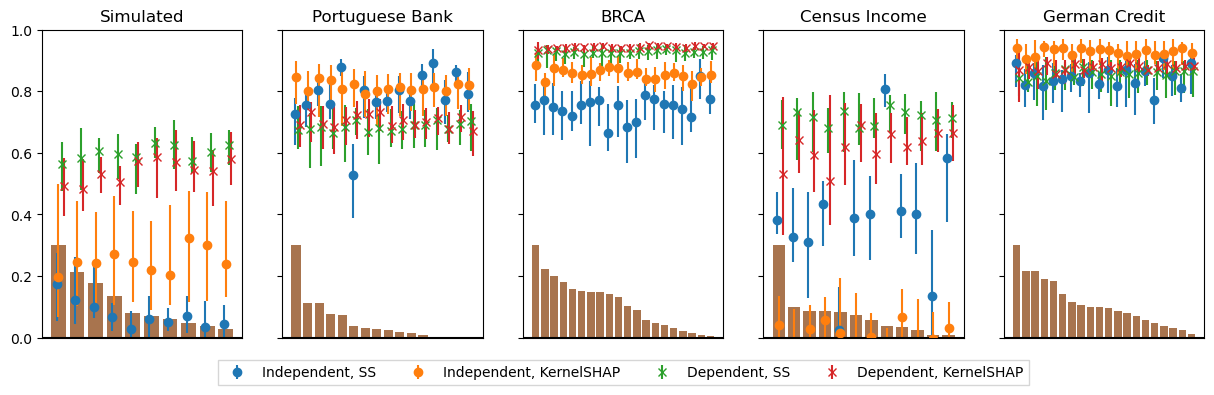}
         \caption{Logistic Regression}
         \label{fig:var reducs glm}
     \end{subfigure}
     \hfill
     \begin{subfigure}[b]{\textwidth}
         \includegraphics[width=\textwidth]{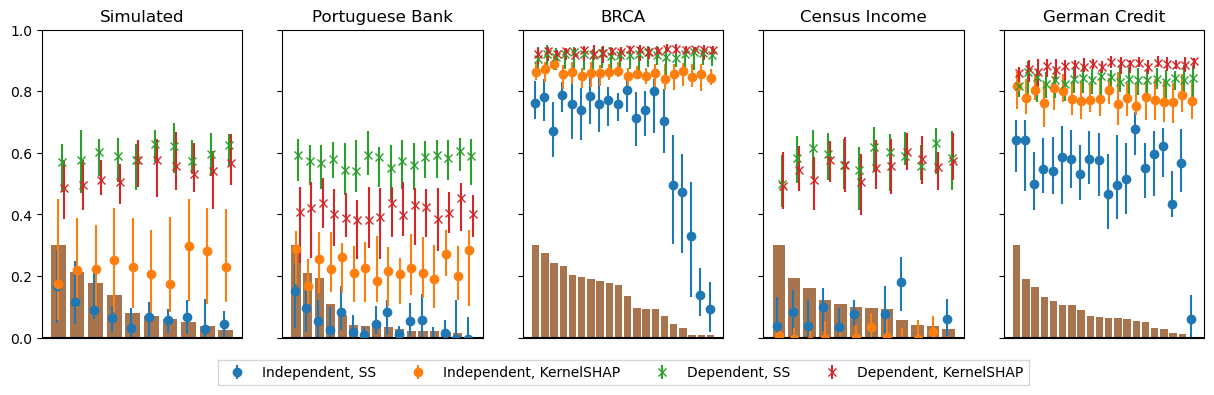}
         \caption{Neural Network}
         \label{fig:var reducs nn}
     \end{subfigure}
     \hfill
     \begin{subfigure}[b]{\textwidth}
         \includegraphics[width=\textwidth]{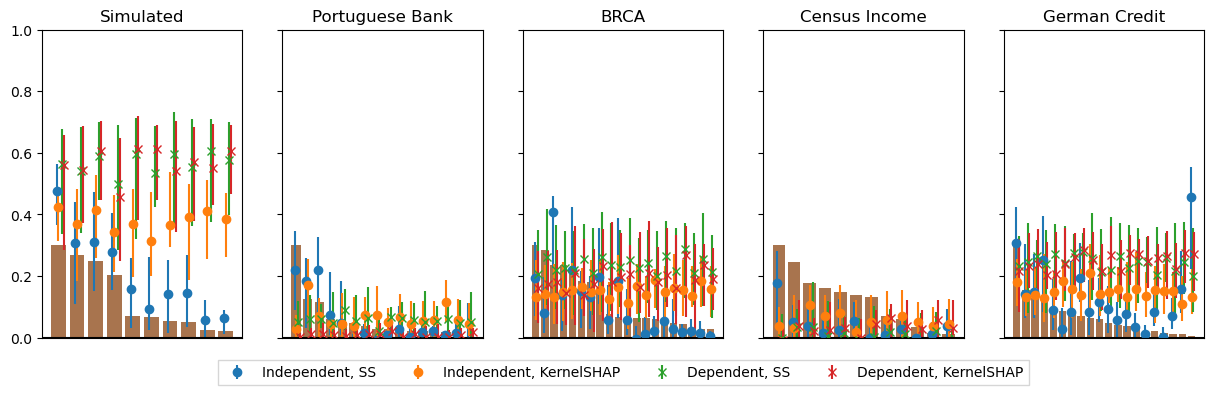}
         \caption{Random Forest}
         \label{fig:var reducs rf}
     \end{subfigure}
        \caption{Variance Reductions}
        \label{fig:var reducs}
\end{figure}

Figure \ref{fig:var reducs} displays the variance reductions of our ControlSHAP methods in the four settings: Independent vs Dependent Features, and Shapley Sampling vs KernelSHAP. The error bars span the 2\textsuperscript{th} to 7\textsuperscript{th} percentiles of the variance reductions for 40 held-out samples. 

\begin{figure}
     \begin{subfigure}[b]{\textwidth}
         \includegraphics[width=\textwidth]{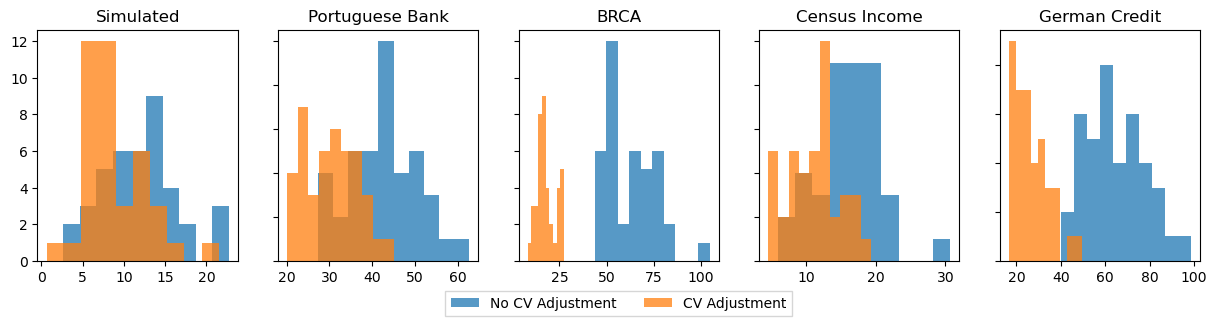}
         \caption{Logistic Regression}
         \label{fig:rank chgs glm}
     \end{subfigure}
     \hfill
     \begin{subfigure}[b]{\textwidth}
         \includegraphics[width=\textwidth]{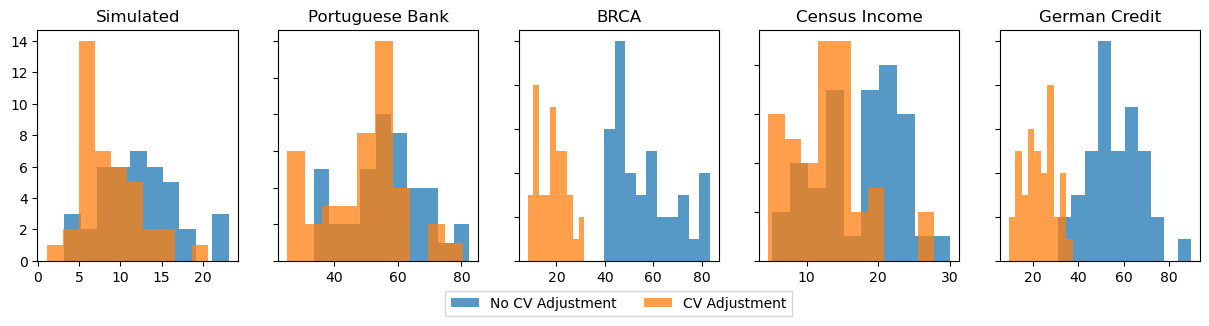}
         \caption{Neural Network}
         \label{fig:rank chgs nn}
     \end{subfigure}
     \hfill
     \begin{subfigure}[b]{\textwidth}
         \includegraphics[width=\textwidth]{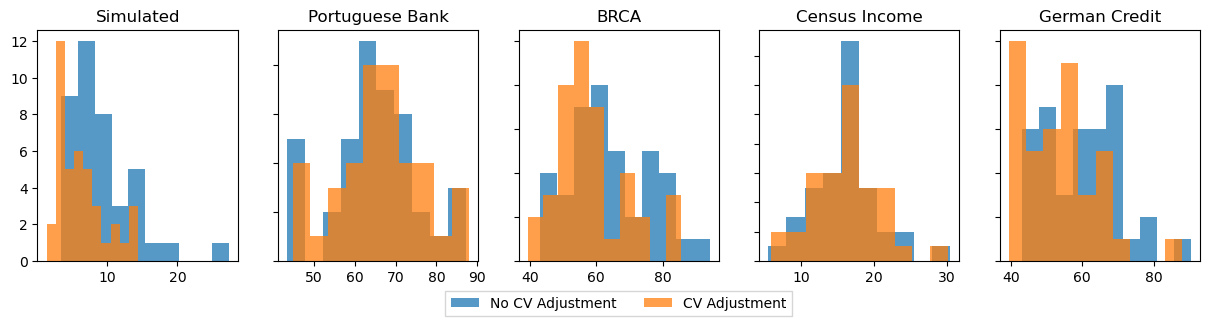}
         \caption{Random Forest}
         \label{fig:rank chgs rf}
     \end{subfigure}
    \centering\caption{Average Number of Rank Changes, with and without ControlSHAP's adjustment via Control Variates (CV).}
    \label{fig:rank chgs}
\end{figure}

Figure \ref{fig:rank chgs} compares the average number of changes in rankings between the original and ControlSHAP Shapley estimates. Specifically, we look the Shapley estimates obtained via KernelSHAP, assuming correlated features.

\section{Anticipated Correlation}

\begin{figure}
    \centering
     \begin{subfigure}[b]{\textwidth}
         \includegraphics[width=\textwidth]{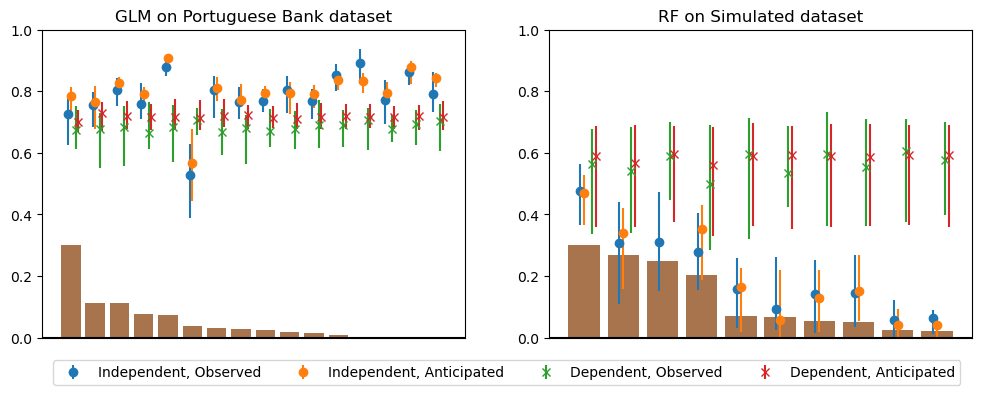}
         \caption{Shapley Sampling}
     \end{subfigure}
     \hfill
     \begin{subfigure}[b]{\textwidth}
         \includegraphics[width=\textwidth]{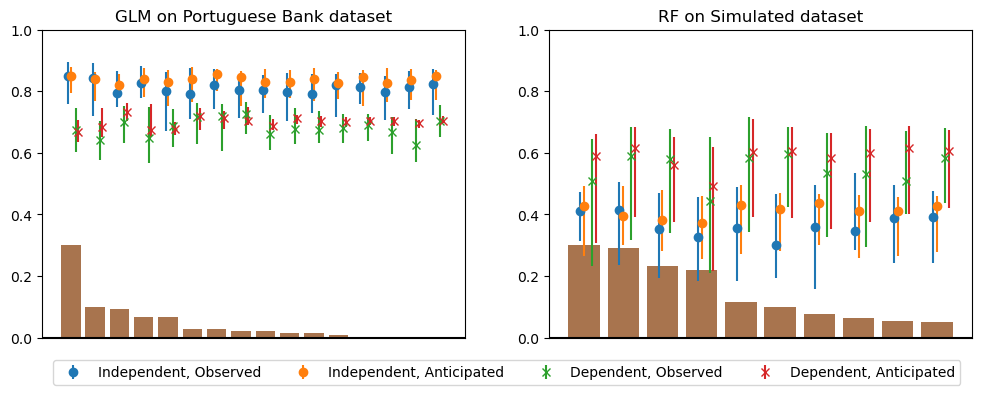}
         \caption{KernelSHAP, Least Squares Variance Estimation}
     \end{subfigure}
     \hfill
     \begin{subfigure}[b]{\textwidth}
         \includegraphics[width=\textwidth]{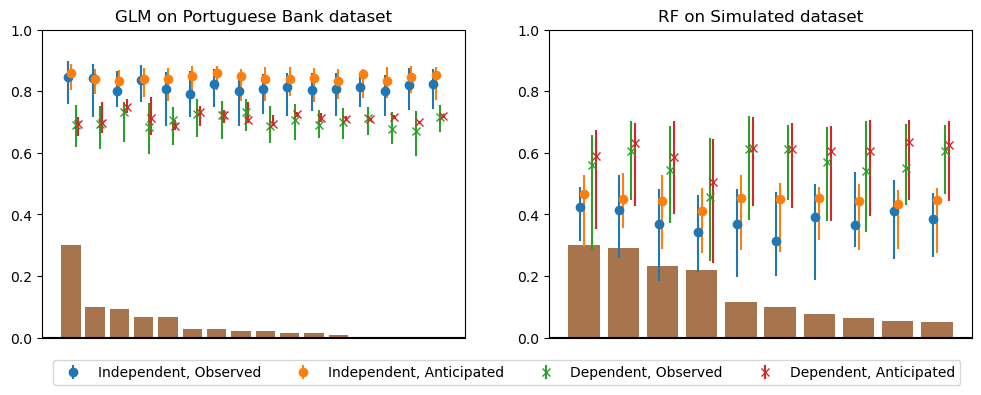}
         \caption{KernelSHAP, Bootstrapped Variance Estimation}
     \end{subfigure}
        \caption{Anticipated vs Observed Variance Reduction}
        \label{fig:anticipated var reducs}
\end{figure}

Figure \ref{fig:anticipated var reducs} compares the observed and \textit{anticipated} variance reductions, across two combinations of dataset and predictor. Recall that the variance reduction of the control variate estimate for $\phi_j(x)$ is $\rho^2(\hat{\phi}_j(x)^\model, \hat{\phi}_j(x)^\apx)$, where $\rho$ is the Pearson's correlation coefficient. We compute the sample correlation coefficient of the Shapley values on sample $x$ as follows:

\begin{equation*}
    \hat{\rho} := \frac{\widehat{\Cov}(\hat{\phi}_j(x)^\model, \hat{\phi}_j(x)^\apx)}{\sqrt{\widehat{\Var}(\hat{\phi}_j(x)^\model)\widehat{\Var}(\hat{\phi}_j(x)^\apx)}}
\end{equation*}

We average this across the 50 iterations to obtain a single estimate for the correlation $\hat{\rho}$. The plots display the median and error bars for $\hat{\rho}^2(\hat{\phi}_j(x)^\model, \hat{\phi}_j(x)^\apx)$ across 40 samples. Once again, the error bars span the 2\textsuperscript{th} to 7\textsuperscript{th} percentiles. 

\section{Comparing KernelSHAP Variance Estimators}

\begin{figure}
    \centering\includegraphics[width=.9\textwidth]{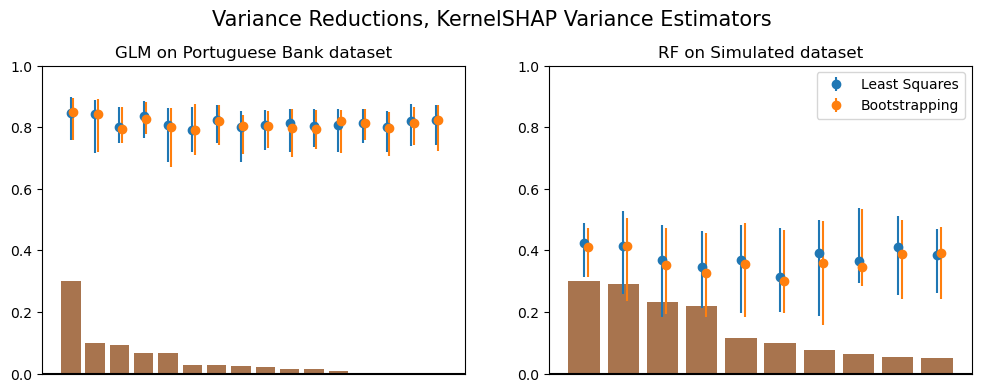}
    \caption{Variance reductions of ControlSHAP across 40 samples with bootstrapped and least-squares estimates of KernelSHAP variance and covariance. Shapley estimates computed assuming independent features.}
    \label{fig:boot vs ls}
\end{figure}

Section 4.2 of the paper details methods for computing the variance and covariance between KernelSHAP estimates. Bootstrapping and the least squares covariance produce extremely similar estimates. In turn, these produce ControlSHAP estimates that reduce variance by roughly the same amount, as shown in Figure \ref{fig:boot vs ls}. This indicates that both methods are appropriate choices.

\begin{figure}[h]
    \centering\includegraphics[width=0.7\textwidth]{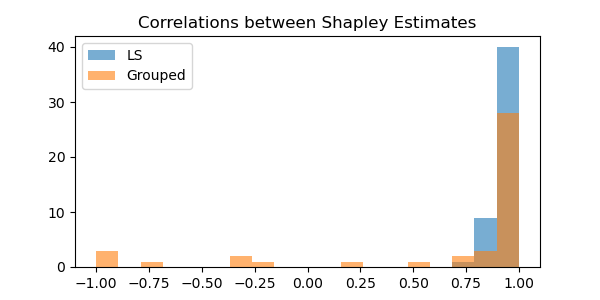}
    \caption{Correlation between model and Taylor approximation, grouped and least squares. Same input and feature across 50 iterations, with logistic regression on the bank dataset.}
    \label{fig:grouped bad}
\end{figure}

In contrast, we were not able to get the grouped method to reliably work. Its variance estimates were somewhat erratic, as they are drawn from a heavy-tailed $\chi^2$ distribution (Figure \ref{fig:grouped bad}). As a result, its ControlSHAP estimates were occasionally  \textit{more} variable than the original KernelSHAP values themselves.



\end{document}